\documentclass[letterpaper, 10 pt, conference]{ieeeconf}  

\usepackage{times}
\usepackage{epsfig}
\usepackage{graphicx}
\usepackage{amsmath}
\usepackage{amssymb}
\usepackage{subcaption}
\usepackage{booktabs}
\usepackage[colorlinks = true,
            linkcolor = magenta,
            urlcolor  = magenta,
            citecolor = blue,
            anchorcolor = blue]{hyperref}
            \usepackage[symbol]{footmisc}

\usepackage{fancyref}
\IEEEoverridecommandlockouts                              

\title{\LARGE \bf
Trajectory Prediction for Autonomous Driving based on Multi-Head Attention with Joint Agent-Map Representation
}

\author{Kaouther Messaoud$^{1,2}$, Nachiket Deo$^{2}$, Mohan M. Trivedi$^{2}$ and Fawzi Nashashibi$^{1}$
\thanks{$^{1}$INRIA Paris,
        {\tt\scriptsize \{kaouther.messaoud,fawzi.nashashibi\}@inria.fr}}%
\thanks{$^{2}$ LISA, UCSD,
        {\tt\scriptsize \{mkaouther,ndeo,mtrivedi\}@ucsd.edu}}%
}

\begin{document}

\maketitle
\thispagestyle{empty}
\pagestyle{empty}

\begin{abstract}


Predicting the trajectories of surrounding agents is an essential ability 
for autonomous vehicles navigating through complex traffic scenes.
The future trajectories of agents can be inferred using two important cues: the locations and past motion of agents, and the static scene structure. Due to the high variability in scene structure and agent configurations, prior work has employed the attention mechanism, applied separately to the scene and agent configuration to learn the most salient parts of both cues. However, the two cues are tightly linked. The agent configuration can inform what part of the scene is most relevant to prediction. The static scene in turn can help determine the relative influence of agents on each other’s motion. Moreover, the distribution of future trajectories is multimodal, with modes corresponding to the agent's intent. The agent's intent also informs what part of the scene and agent configuration is relevant to prediction.
We thus propose a novel approach applying multi-head attention by considering a joint representation of the static scene and surrounding agents. We use each attention head to generate a distinct future trajectory to address multimodality of future trajectories. Our model achieves state of the art results on the nuScenes prediction benchmark and generates diverse future trajectories compliant with scene structure and agent configuration.

\end{abstract}

\section{Introduction}\label{sec:intro}

Autonomous vehicles navigate in highly-uncertain and interactive environments shared with other dynamic agents. In order to plan safe and comfortable maneuvers, they need to predict future trajectories of surrounding vehicles. The inherent uncertainty of the future makes trajectory prediction a challenging task. However, there is structure to vehicle motion. Two cues in particular provide useful context to predict the future trajectories of vehicles: (1) The past motion of the vehicle of interest, and the motion of its neighbouring agents and (2) the static scene structure including road and lane structure, sidewalks and crosswalks. 

A major challenge in trajectory prediction is the high variability in both context cues. Static scene elements and agents in a traffic scene can have various configurations. Several existing machine learning models for trajectory prediction have employed the attention mechanism with single \cite{AVemula,Sadeghian,Park} or multiple heads \cite{KMessaoud, KMessaoud2, messaoud3} to learn the most salient subsets of the static scene and agent configuration, to make both cues tractable. However, a limitation of existing approaches is that they either consider just one of the two cues \cite{AVemula, KMessaoud, KMessaoud2}, or apply attention modules separately to representations of the agent configuration and static scene \cite{Sadeghian}. The two cues are tightly linked; each can inform the most salient parts of the other. For example, the presence of a nearby pedestrian could make a crosswalk in the path of the vehicle of interest a salient part of the static scene, as opposed to if there were no nearby pedestrians. On the other hand, the presence of a crosswalk would make a nearby pedestrian on the sidewalk a more salient part of the input context as opposed to if there was no crosswalk.

Another challenge in trajectory prediction is the multimodality of the distribution of future trajectories. The vehicle of interest could have one of several plausible intents, each of which would correspond to a distinct future trajectory. To address multimodality most existing approaches build one latent representation of the context~\cite{NDeo,NDeo3,Gupta, Zhao} and then generate multiple possible trajectories based on this representation. However, we believe that each possible future trajectory is conditioned on a specific subset of surrounding agents' behaviors and scene context features. For each possible intent, a different partial context is important to understand the future behavior.

To address the above challenges, we propose a model for multimodal trajectory prediction of vehicles, utilizing multi-head attention as proposed in \cite{AVaswani}. In particular, our model has the following characteristics:

\begin{enumerate}
    \item \textbf{Joint agent-map representation:} Unlike prior approaches, we use a joint representation of the agents and static scene (represented as an HD map) to generate keys and values for attention heads. This allows us to better model the inter-dependency of both cues. Our experiments show that attention heads applied to a joint agent-map representation outperform those applied to separate representations of agents and the map. 
    \item \textbf{Attention heads specializing in prediction modes:} We  model the predictive distribution as a mixture model, with each attention head specializing in one mixture component. This allows each attention head to weight different parts of the scene and agent configuration conditioned on agent intent. Our experiments show that attention heads specializing in modes of the predictive distribution outperform an ensemble of attention heads used for predicting every mode.  
\end{enumerate}

We evaluate our model on the publicly available NuScenes dataset \cite{nuscenes2019}. Our model achieves state of the art results on the NuScenes prediction benchmark, outperforming all entries on 9 of the 11 evaluation metrics. Our model generates diverse future trajectories, that conform to the static scene and agent configuration. The code for the proposed model will be made available at: \href{https://github.com/KaoutherMessaoud/MHA-JAM}{https://github.com/KaoutherMessaoud/MHA-JAM}.

\begin{figure*}[t]
\begin{center}
   \includegraphics[width=0.8\linewidth]{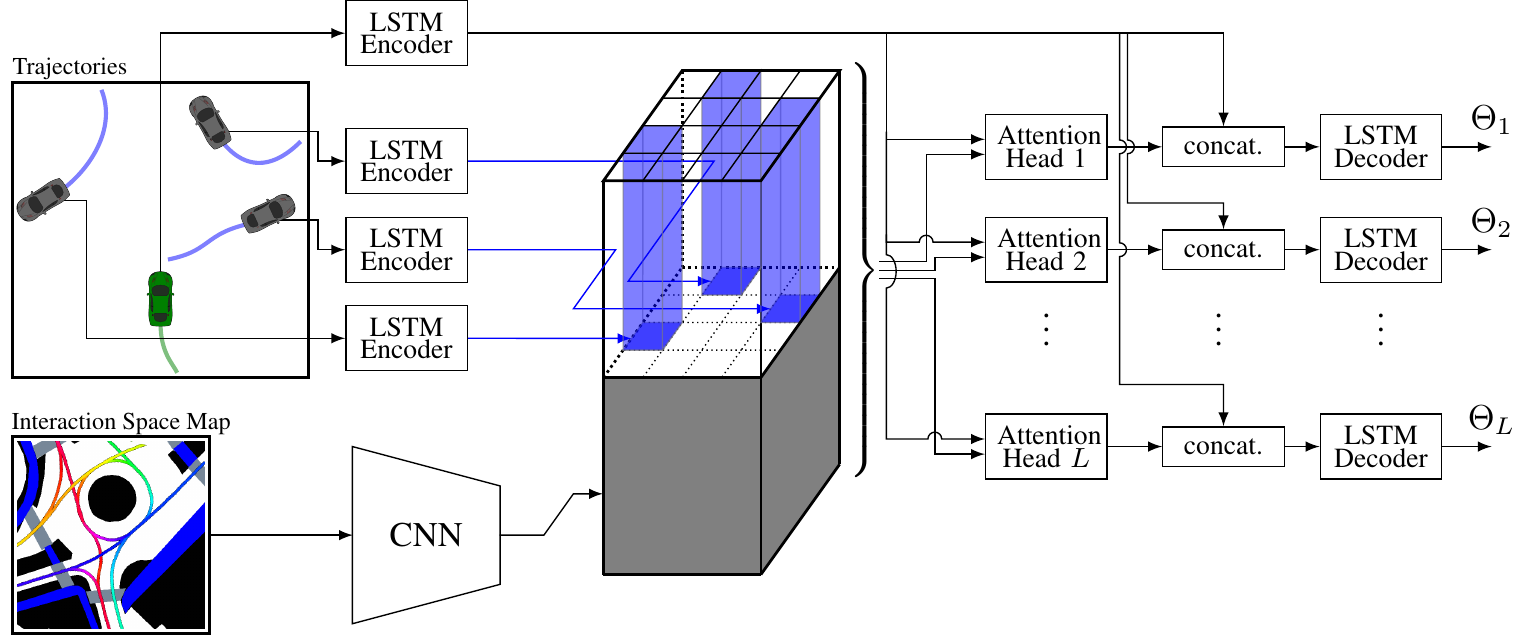}
\end{center}
   \caption{\textbf{MHA-JAM} (MHA with Joint Agent Map representation): Each LSTM encoder generates an encoding vector of one of the surrounding agent recent motion. The CNN backbone transforms the input map image to a 3D tensor of scene features. A combined representation of the context is build by concatenating the surrounding agents motion encodings and the scene features. Each attention head models a possible way of interaction between the target (green car) and the combined context features. Each LSTM decoder receives an context vector and the target vehicle encoding and generates a possible distribution over a possible predicted trajectory conditioned on each context.}
\label{fig:MHA}
\end{figure*}

\section{Related Research}\label{sec:related_work}

\noindent\textbf{Cross-agent interaction:}
Drivers and pedestrians operate in a shared space, co-operating with each other to perform safe maneuvers and reach their goals.  Thus, most state of the art trajectory prediction approaches model cross-agent interaction. Alahi \textit{et al.,}~\cite{AAlahi} propose \textit{social pooling}, where the LSTM states of neighboring agents were pooled based on their locations in a 2-D grid to form \textit{social tensors} as inputs to the prediction model. Deo \textit{et al.}~\cite{NDeo} extend this concept to model more distant interactions using successive convolutional layers. Messaoud \textit{et al.,}~\cite{KMessaoud} instead apply a multi-head attention mechanism to the social tensor to directly relate distant vehicles and extract a context representation. Our approach is closest to ~\cite{KMessaoud}, however, we additionally consider an encoding of the static scene as an input to the multi-head attention modules. 

\vspace{0.05in}

\noindent\textbf{Agent-scene modeling:}
Static scene context in the bird's eye view is also an important cue that has been exploited in prior work for trajectory prediction. Zhao \textit{et al.}~\cite{Zhao} concatenate social tensors and an encoding of the scene in the bird's eye view and apply convolutional layers to extract a joint representation of the scene and surrounding agent motion. Sadeghian \textit{et al.}~\cite{Sadeghian} deployed two parallel attention blocks; a social attention for vehicle-vehicle interactions and a physical attention for vehicle-map interactions modeling. Yuan \textit{et al.}~\cite{Park} use two attention modules as well but deploy them sequentially by feeding the output of the cross-agent attention module as a query to the scene attention module. Unlike ~\cite{Park, Sadeghian}, our model generates a joint representation of the scene and social tensor similar to ~\cite{Zhao}, which we use as an input to multi-head attention modules to model the most salient parts of the scene and agent configuration.

\vspace{0.05in}

\noindent\textbf{Multimodal trajectory prediction:} 
The future motion of agents is inherently multimodal, with multiple plausible trajectories conditioned on the agents' intent. Thus recent approaches learn one to many mappings from input context to multiple future trajectories. The most approach is sampling generative models such as conditional variational autoencoder (CVAE)~\cite{Desire} and Generative Adversarial Networks (GANs)~\cite{Zhao,Gupta,Sadeghian}. Other methods sample a stochastic policies learnt by imitation or inverse reinforcement learning~\cite{Li_2019_CVPR,Deo2020TrajectoryFI}.
Ridel \textit{et al.}~\cite{DRidel} predict the probability distributions over grids and generates multiple trajectory samples. In this paper, we utilise a mixture model similar to ~\cite{HCui, Multipath}, generating a fixed number $L$ of plausible trajectories for an agent, where each
trajectory is represented by a sequence of two-dimensional Gaussians and the probability associated with each of the $L$ trajectories.

\section{Multi-head Attention with \\ Joint Agent-Map Representation}\label{sec:model}

\subsection{Input Representation}



\noindent \textbf{Interaction space:} We first define the \textit{interaction space} of a target vehicle $T$ as the area centered on its position at the prediction instant $t_{pred}$ and oriented in its direction of motion, and denote it as $\mathcal{A}_{T}$. We consider all agents present in $\mathcal{A}_{T}$ and the static scene elements in it as inputs to our model. This representation enables us to consider a varying number of interacting agents based on their occupancy in this area. We consider $\mathcal{A}_{T}$ to extend $\pm 25 m$ from the target vehicle $T$ in the lateral direction, $40 m$ in the longitudinal direction ahead of $T$ and $10 m$ behind it.

\vspace{0.05in}

\noindent \textbf{Trajectory representation:}
Each agent $i$ in the interaction space is represented by a sequence of its states, for $t_h$ past time steps between $t_{pred}-t_h$ and $t_{pred}$,
\begin{equation}
    S_{i}=[S_{i}^{t_{pred}-t_h},\dots,S_{i}^{t_{pred}}].
\end{equation}
Here, superscripts denote time, while subscripts denote agent index. Absence of subscripts implies all agents in $\mathcal{A}_{T}$, and absence of superscripts implies all time steps from $t_{pred}- t_h$ to $t_{pred}$. Each state is composed of a sequence of the agent relative coordinates $x_i^t$ and $y_i^t$, velocity $v_i^t$, acceleration $a_{i}^t$ and yaw rate $\dot{\theta}_i^t$,
\begin{equation}
    S_{i}^t =(x_i^t,y_i^t,v_{i}^t,a_{i}^t,\dot{\theta}_i^t).
\end{equation}
The positions are expressed in a stationary frame of reference where the origin is the position of the target vehicle at the prediction time $t_{pred}$. The $y$-axis is oriented toward the target vehicle's direction of motion and $x$-axis points to the direction perpendicular to it.

\vspace{0.05in}

\noindent \textbf{Map representation:} 
We use a rasterized bird's eye view map similar to ~\cite{HCui}, to represent the static scene elements in $\mathcal{A}_{T}$. The map includes the road geometry, drivable area, lane structure and direction of motion along each lane, locations of sidewalks and crosswalks. We denote the map as $\mathcal{M}$.


\subsection{Multimodal Output Representation}
We wish to estimate the probability distribution $P(Y|S,\mathcal{M})$ over the future locations $Y$ of the target vehicle, conditioned on the past trajectories $S$ of agents in $\mathcal{A}_{T}$, and the map $\mathcal{M}$. To account for multimodality of $P(Y|S,\mathcal{M})$, we model it as a mixture distribution with $L$ mixture components. Each mixture component consists of predicted location co-ordinates at discrete time-steps over a prediction horizon $t_f$.   
\begin{equation}
    Y_l=[Y_l^{t_{pred}+1},\dots, Y_l^{t_{pred}+t_{f}}], ~l=1,\dots, L.
\end{equation}

Here, superscripts denote time, while subscripts denote the mixture component. Each predicted location $Y_l^t$ is modeled as a bivariate Gaussian distribution. Our model outputs the means and variances $\Theta^t_l=(\mu^t_l,\Sigma^t_l)$ of the Gaussian distributions for each mixture component at each time step. Additionally it outputs the probabilities $P_l$ associated with each mixture component.

\subsection{Encoding layers}

The encoding layers generate feature representations of the past trajectories $S$, and the map $\mathcal{M}$. They comprise:

\vspace{0.05in}

\noindent \textbf{Trajectory encoder:}
The state vector $S_{i}^{t}$ of each agent is embedded using a fully connected layer to a vector $e_{i}^{t}$ and encoded using an LSTM encoder from $t_{pred}- t_h$ to $t_{pred}$,
\begin{equation}
h_{i}^{t}=LSTM(h_{i}^{t-1},e_{i}^{t};W_{enc}).
\end{equation}

 Here, $h_{i}^{t}$ and $h_{T}^{t}$ are the hidden states vector of the $i^{th}$ surrounding agent and the target vehicle respectively at time $t$. All the LSTM encoders share the same weights $W_{enc}$.
 
 \vspace{0.05in}

\noindent\textbf{Map encoder:} We use a CNN to extract high level features from the map $\mathcal{M}$. The CNN outputs a map feature representation $\mathcal{F}_m$ of size $(M,N,C_m)$, where $(M,N)$ represent spatial dimensions of the feature maps and $C_m$ the number of feature channels.
\subsection{Joint Agent-Map Attention}\label{MHA-JAM}
The first step in modeling vehicle-agents and vehicle-map interactions is to build a combined representation of the global context. To do so, we first build a \textit{social tensor} as proposed in \cite{AAlahi}. We divide $\mathcal{A}_T$ into a spatial grid of size $(M,N)$. The trajectory encoder states of the surrounding agents $h_{i}^{t_{pred}}$ at the prediction instance are placed at their corresponding positions in the 2D spatial grid, giving us a tensor $\mathcal{F}_s$ of size $(M,N, C_{h})$, where $C_h$ is the size of the trajectory encoder state. Figure \ref{fig:MHA} shows an example of a social tensor. We then concatenate the social tensor $\mathcal{F}_s$ and map features $\mathcal{F}_m$ along the channel dimension to generate a combined representation $\mathcal{F}$ of agent trajectories and the map,
\begin{equation}
   \mathcal{F}= Concat(\mathcal{F}_s,\mathcal{F}_m).
\end{equation}
\begin{figure}[t]
\centering
\includegraphics[width=\columnwidth]{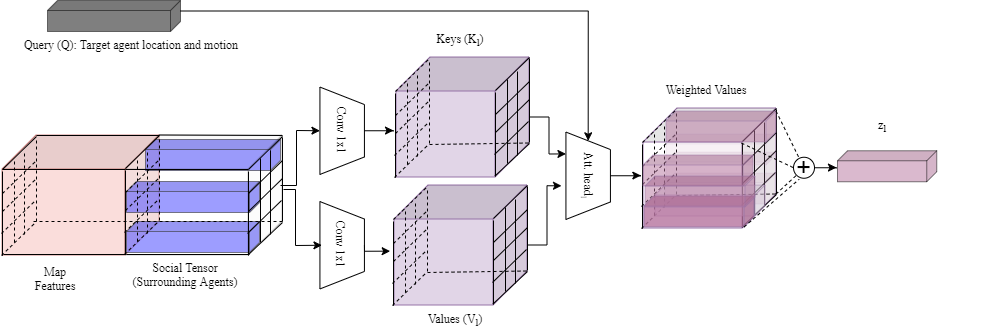}
\caption{\textbf{Attention modules in MHA-JAM:} We generate keys and values by applying 1x1 convolutional layers to a joint representation of the map and surrounding agents, while the trajectory encoding of the target agent serves as the query. }
\label{fig:JAM}
\end{figure}
We use the multi-head attention mechanism \cite{AVaswani} to
extract the salient parts of the joint agent-map representation $\mathcal{F}$ as shown in figure \ref{fig:JAM}, with an attention head assigned to each of the $L$ mixture components. For each attention head, the hidden state of the target vehicle $h_T^{t_{~pred}}$ is processed by a fully connected layer $\theta_l$ to give the query  
\begin{equation}
Q_l=\theta_l(h_T^{t_{~pred}};W_{\theta_l}).
\end{equation}
The agent-map representation $\mathcal{F}$ is processed by $1\times1$ conv. layers $\phi_l$ and $\rho_l$ to give keys $K_l$ and values $V_l$, 
\begin{equation}
K_l=\phi_l (\mathcal{F};W_{\phi_l}), 
\end{equation}
\begin{equation}
V_l=\rho_l (\mathcal{F};W_{\rho_l}). 
\end{equation}

The output $A_l$ of each attention head is then calculated as a weighted sum of value vectors $V_{l}(m,n,:)$,

\begin{equation}
A_l=\sum_{m=1}^{M}\sum_{n=1}^{N}{\alpha_{l}(m,n){V_{l}(m,n,:)}}.
\end{equation}

Here, $\alpha_{l}(m,n)$ weights the effect of each value vector $V_l(m,n)$,
\begin{equation}
\alpha_l(m,n) =Softmax(\frac{Q_lK_l^T(m,n,:)}{\sqrt{d}}),
\end{equation}

where $Q_lK^T_l(m,n,:)$ is the dot product of  $Q_l$ and $K_l(m,n,:)$ and $d$ is the dimension of $Q_l$ and $K_l(m,n,:)$.

For each attention head we concatenate the output $A_l$
with the target vehicle trajectory encoder state $h_{T}^{t_{pred}}$ to give a context representation $z_l$ for each mixture component $l=1,\dots, L$,
\begin{equation}
z_l=Concat(h_{T}^{t_{pred}}, A_l).
\end{equation}
\subsection{Decoding layer}
Each context vector $z_l$, representing the selected information about the target vehicle's interactions with the surrounding agents and the scene, and its motion encoding are fed to $l$ LSTM Decoders.
The decoders generate the predicted parameters of the distributions over the target vehicle's estimated future positions of each possible trajectory for next $t_{f}$ time steps,
\begin{equation}
\Theta^{t}_l=\Lambda (LSTM(h_{l}^{t-1},z_l;W_{dec})).
\end{equation}
All the LSTM decoders share the same weights $W_{dec}$ and $\Lambda$ is a fully connected layer.
Similar to ~\cite{HCui}, we also output the probability $P_l$ associated with each mixture component. To do so, we concatenate all the scene representation vectors $z_l$, feed them to two successive fully connected layers and apply the softmax activation to obtain L probability values.

\subsection{Loss Functions}\label{lossfn}
\vspace{0.05in}
\noindent \textbf{Regression loss}:
While the model outputs a multimodal predictive distribution corresponding to L distinct futures, we only have access to 1 ground truth trajectory for training the model. In order to not penalize plausible trajectories generated by the model that do not correspond to the ground truth, we use a variant of the best of L regression loss for training our model, as has been previously done in \cite{Gupta}. This encourages the model to generate a diverse set of predicted trajectories. Since we output the parameters of a bivariate Gaussian distribution at each time step for the $L$ 
trajectories, we compute the negative log-likelihood (NLL) of the ground truth trajectory under each of the $L$ modes output by the model, and consider the minimum of the $L$ NLL values as the regression loss. The regression loss is given by
\begin{equation}
\label{eq:nll}
L_{reg}=\min_{l}\sum_{t=t_{pred+1}}^{t_{pred}+t_{f}} - log ( P_{\Theta^{t}_l} (Y_l^{t}|S, \mathcal{M})). 
\end{equation}

\vspace{0.1in}

\noindent\textbf{Classification loss }:
In addition to the regression loss, we consider the cross entropy as used in ~\cite{HCui, Multipath}, 
\begin{equation}
\label{eq:class}
L_{cl} = -\sum_{l=1}^{L}\delta _{l^{*}}(l)log(P_l),
\end{equation}
where $\delta$ is a function equal to 1 if $l=l^{*}$ and 0 otherwise. Here $l^{*}$ is the mode corresponding to the minimum NLL in equation \ref{eq:nll}. $Y_{l^*}$ is the predicted trajectory corresponding to $l^{*}$ and $P_l^*$ its predicted probability.

\vspace{0.1in}

\noindent\textbf{Off-road loss}:
While the loss given by equation \ref{eq:nll} encourages the model to generate a diverse set of trajectories, we wish to generate trajectories that conform to the road structure. Since the regression loss only affects the trajectory closest to the ground-truth, we consider the auxiliary loss function proposed in \cite{niedoba, Park} that penalizes points in any of the L trajectories that lie off the road. The off-road loss $L_{or}$ for each predicted location is the minimum distance of that location from the drivable area.\\
The overall loss for training the model is given by 
\begin{equation}
    L = L_{reg} + \lambda_{cl}.L_{cl} + \lambda_{or}.L_{or},
\end{equation}

where the the weights $\lambda_{cl}$ and $\lambda_{or}$ are empirically determined hyperparameters.


\subsection{Implementation details}
The input states are embedded in a space of dimension 32. We use an image representation of the scene map of size of $(500,500)$ with a resolution of $0.1$ meters per pixel. Similar to~\cite{Covernet} representation, our input image extents are $40~m$ ahead of the target vehicle, $10~m$ behind and $25~m$ on each side. We use ResNet-$50$ pretrained on ImageNet to extract map features. This CNN outputs a map features of size $(28,28, 512)$ on top of them we place the trajectories encodings.
The deployed LSTM encoder and decoder are of 64 and 128 randomly initialized units respectively. We use $L=16$ parallel attention operations applied on the vectors projected on different spaces of size d=64. We use a batch size of 32 and Adam optimizer~\cite{DPKingma}. The model is implemented using PyTorch~\cite{APaszke}.

\section{Experimental Analysis and Evaluations}\label{sec:results}



\subsection{Dataset}
We train and evaluate our model using the publicly available nuScenes~\cite{nuscenes2019} dataset. The dataset was captured using vehicle mounted camera and lidar sensors driving through Boston and Singapore. It comprises 1000 \textit{scenes}, each of which is a 20 second record, capturing complex inner city traffic. Each scene includes agent detection boxes and tracks hand-annotated at 2 Hz, as well as high definition maps of the scenes. We train and evaluate our model using the official benchmark split for the nuScenes prediction challenge, with 32,186 prediction instances in the train set, 8,560 instances in the validation set, and 9,041 instances in the test set.

\begin{table*}
\caption{Results of comparative analysis on nuScenes dataset, over a prediction horizon of 6-seconds}
\centering
\resizebox{\textwidth}{!}{%
\begin{tabular}{lccccccccccc}\toprule
 & MinADE$_1$& MinADE$_5$ & MinADE$_{10}$ & MinADE$_{15}$& MinFDE$_1$ & MinFDE$_5$ & MinFDE$_{10}$ & MinFDE$_{15}$ & MissRate$_{5,2}$ & MissRate$_{10,2}$ & Off-Road Rate \\\midrule
Const vel and yaw & 4.61    & 4.61     & 4.61  & 4.61  & 11.21   & 11.21    & 11.21  & 11.21   & 0.91      & 0.91    & 0.14 \\
Physics oracle    & \textbf{3.69}    & 3.69     & 3.69  & 3.69  & 9.06    & 9.06     & 9.06   & 9.06   & 0.88 & 0.88          & 0.12 \\
MTP~\cite{HCui}               & 4.42    & 2.22     & 1.74  & 1.55  & 10.36   & 4.83     & 3.54   & 3.05   & 0.74  & 0.67       & 0.25 \\
Multipath~\cite{Multipath}               & 4.43    & \textbf{1.78}    & 1.55  & 1.52  & 10.16   & \textbf{3.62}     & 2.93   & 2.89   & 0.78   & 0.76       & 0.36 \\
CoverNet\footnotemark[1]{}~\cite{Covernet}               &  -   & 2.62    & 1.92  & -  & 11.36   & -     & -   & -   & 0.76   & 0.64       & 0.13\\
Trajectron$++$\footnotemark[1]{}~\cite{Trajectron}               & - & 1.88  & 1.51 & -  &  9.52  &  -     & -   & -   & 0.70   & 0.57       & 0.25\\\midrule
MHA-JAM & 3.77    & 1.85     & \textbf{1.24}  & \textbf{1.03}  & 8.65    & 3.85     & 2.23  & \textbf{1.67}    & 0.60         & 0.46 & 0.10\\
MHA-JAM (off-road)\footnotemark[1]{}  & \textbf{3.69}    & 1.81     & \textbf{1.24}  & \textbf{1.03}  & \textbf{8.57}    & 3.72     & \textbf{2.21}  & 1.70    & \textbf{0.59}          & \textbf{0.45} & \textbf{0.07}\\
\bottomrule
\end{tabular}%
}
\label{tab:tab1}
\end{table*}
\subsection{Baselines}
We compare our model to six baselines, two physics based approaches, and four recently proposed models that represent the state of the art for multimodal trajectory prediction. All deep learning based models generate up to $L = 25$ trajectories ($L = 16$ for all implemented methods) and their likelihoods. We report the results considering the $k$ most probable trajectories generated by each model.

\vspace{0.05in}
\noindent\textbf{Constant velocity and yaw :} Our simplest baseline is a physics based model that computes the future trajectory while maintaining constant velocity and yaw of the current state of the vehicle.

\vspace{0.05in}
\noindent\textbf{Physics oracle :} An extension of the physics based model introduced in~\cite{Covernet}. Based on the current state of the vehicle (velocity, acceleration and yaw), it computes the minimum average point-wise Euclidean distance over the predictions generated by four models: (i) constant velocity and yaw, (ii) constant velocity and yaw rate, (iii) constant acceleration and yaw, and (iv) constant acceleration and yaw rate.

\vspace{0.05in}
\noindent\textbf{Multiple-Trajectory Prediction (MTP)}~\cite{HCui}: The MTP model uses a CNN over a rasterized representation of the scene and the target vehicle state to generate a fixed number of trajectories (modes) and their associated probabilities. It uses a weighted sum of regression (cf. Equation\ref{eq:nll}) and classification (cf. Equation\ref{eq:class}) losses for training. We use the implementation of this model by~\cite{Covernet}.

\vspace{0.05in}
\noindent\textbf{Multipath}~\cite{Multipath}:
Similar to MTP, the Multipath model uses a CNN with same input. However, unlike MTP, it uses fixed \textit{anchors} obtained from the train set to represent the modes, and outputs residuals with respect to anchors in its regression heads. We implement MultiPath as described in ~\cite{Multipath}.


\vspace{0.05in}
\noindent\textbf{CoverNet}~\cite{Covernet}: 
The CoverNet model formulates multimodal trajectory prediction purely as a classification problem. The model predicts the likelihood of a fixed trajectory set, conditioned on the target vehicle state.  

\vspace{0.05in}
\noindent\textbf{Trajectron$++$}~\cite{Trajectron}: is a graph-structured recurrent model that predicts the agents trajectories while considering agent motions and heterogeneous scene data.
\footnotetext[1]{\label{note1} Results reported in the nuScenes challenge Leaderboard:\\ \url{https://evalai.cloudcv.org/web/challenges/challenge-page/591/leaderboard/1659}}

\subsection{Metrics}

\noindent\textbf{$\mbox{MinADE}_k$ and $\mbox{MinFDE}_k$}:
We report the minimum average and final displacement errors over $k$ most probable trajectories similar to prior approaches for multimodal trajectory prediction ~\cite{Gupta, Desire, HCui, Multipath, Deo2020TrajectoryFI}. The minimum over $k$ avoids penalizing the model for generating plausible future trajectories that don't correspond to the ground truth.



%

\vspace{0.05in}
\noindent\textbf{Miss rate:} For a given distance $d$, and the $k$ most probable predictions generated by the model, the set of $k$ predictions is considered a miss based on, 
\begin{equation}
\mbox{Miss}_{k,d}=\left\{\begin{smallmatrix}
1 & \scalebox{0.8}{if }\min_{\hat{\mathbf{y}}\in P_k}\left(\max_{t=t_{pred}}^{t=t_{f}}\left \| \mathbf{y}^t - \hat{\mathbf{y}}^t \right \|\right) \geqslant d.\\
0 & \scalebox{0.8}{otherwise}\\
\end{smallmatrix}\right.
\end{equation}
The miss rate MissRate$_{k,d}$ computes the fraction of missed predictions over the test set.

\vspace{0.05in}
\noindent\textbf{Off-road rate:} Similar to \cite{Deo2020TrajectoryFI}, we consider the off-road rate, which measures the fraction of predicted trajectories that fall outside the drivable area of the map.

\subsection{Quantitative Results}

We compare our model \textbf{MHA-JAM} (MHA with joint agent map representation trained with off road loss) to various baselines in table \ref{tab:tab1}. Our model outperforms all baselines on 9 of the 11 reported metrics, while being second on the remaining two, representing the state of the art on the nuScenes benchmark as of writing this paper. 

For the MinADE$_{k}$ and MinFDE$_{k}$ metrics, our model achieves the best results for $k \in \left \{ 1,10,15 \right \}$ and second best to Multipath \cite{Multipath} when $k = 5$. Having the best performance for $k\in\{10,15\}$ shows that our method generates a diverse set of plausible trajectories that match the ground truth. However, for $k=5$, our classifier doesn't seem to succeed in selecting the closest trajectories to the ground truth among the $5$ most probable ones, while the Multipath classifier does.

Moreover, our method presents significant improvements compared to others when considering miss rate and off-road rate metrics. Having the lowest miss rate suggests that our predicted trajectories are less likely to deviate from the ground truth over a threshold of $d=2m$. In addition, our model achieves significantly lower off-road rates especially when trained with the off-road loss that penalizes predictions outside of the drivable area. Therefore, it generates scene compliant trajectories.

\subsection{Ablation Experiments:}
\begin{figure*}[]
\centering
\begin{subfigure}{\textwidth}
\centering
\begin{subfigure}[t]{0.3\textwidth}
\renewcommand\thesubfigure{i}
\includegraphics[width=0.9\linewidth]{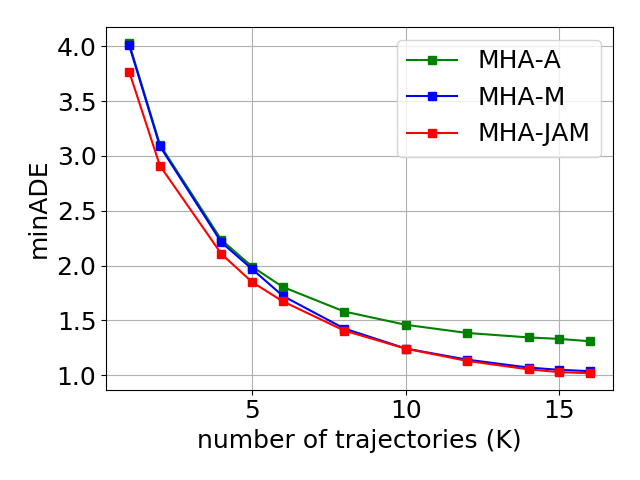}
\end{subfigure}
\begin{subfigure}[t]{0.3\textwidth}
\renewcommand\thesubfigure{i}
\includegraphics[width=0.9\linewidth]{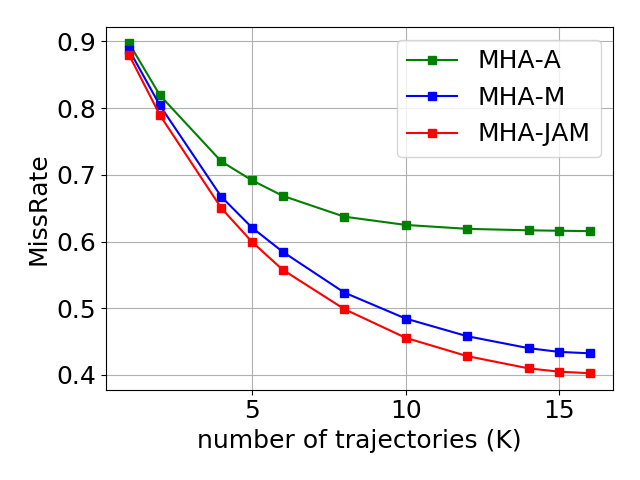}
\end{subfigure}
\begin{subfigure}[t]{0.3\textwidth}
\renewcommand\thesubfigure{i}
\includegraphics[width=0.9\linewidth]{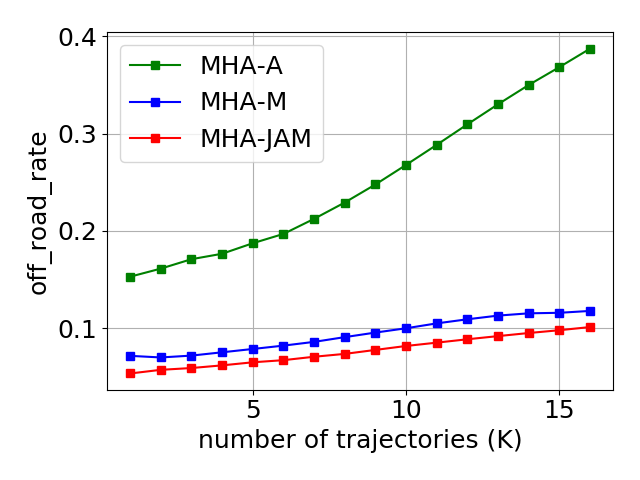}
\end{subfigure}
\caption{Input cues importance: evaluation metrics for different input features.}
\label{fig:Cues}
\end{subfigure}

\begin{subfigure}{\textwidth}
\centering
\begin{subfigure}[t]{0.3\textwidth}
\renewcommand\thesubfigure{i}
\includegraphics[width=0.9\linewidth]{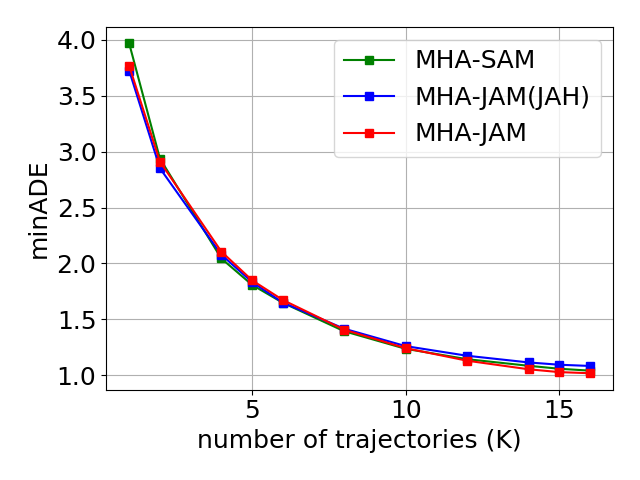}
\end{subfigure}
\begin{subfigure}[t]{0.3\textwidth}
\renewcommand\thesubfigure{i}
\includegraphics[width=0.9\linewidth]{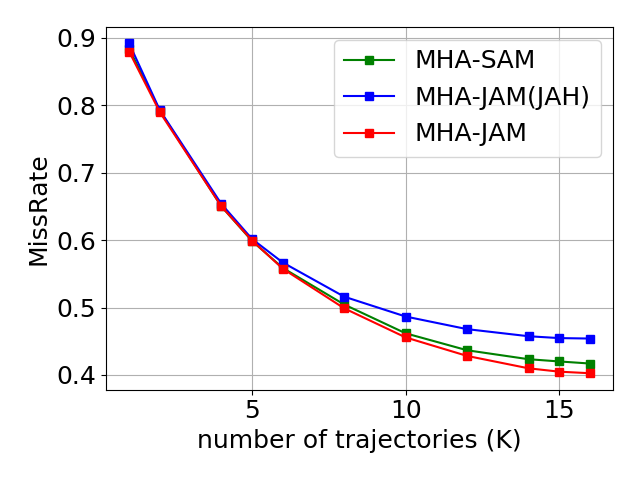}
\end{subfigure}
\begin{subfigure}[t]{0.3\textwidth}
\renewcommand\thesubfigure{i}
\includegraphics[width=0.9\linewidth]{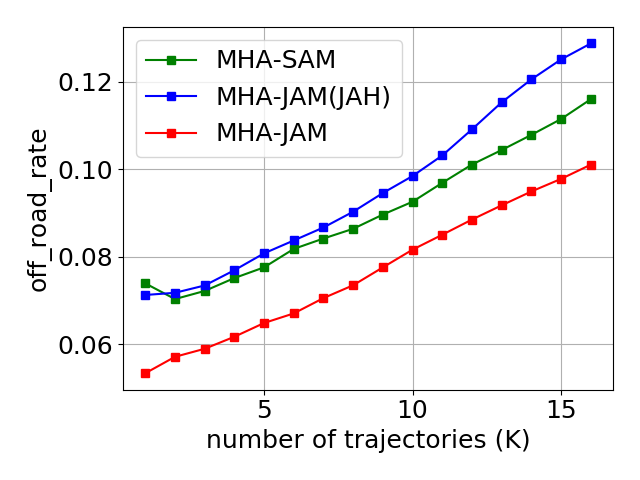}
\end{subfigure}
\caption{Context representation importance: comparison of MHA-SAM, MHA-JAM(JAH) and MHA-JAM}
\label{fig:SAM_JAM}
\end{subfigure}

\begin{subfigure}{\textwidth}
\centering
\begin{subfigure}[t]{0.3\textwidth}
\renewcommand\thesubfigure{i}
\includegraphics[width=0.9\linewidth]{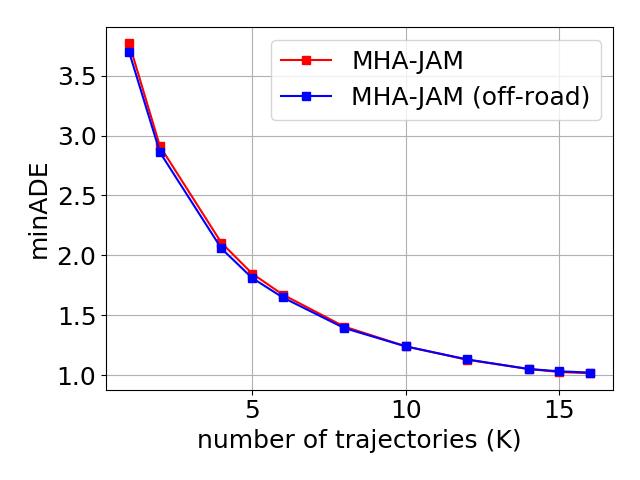}
\end{subfigure}
\begin{subfigure}[t]{0.3\textwidth}
\renewcommand\thesubfigure{i}
\includegraphics[width=0.9\linewidth]{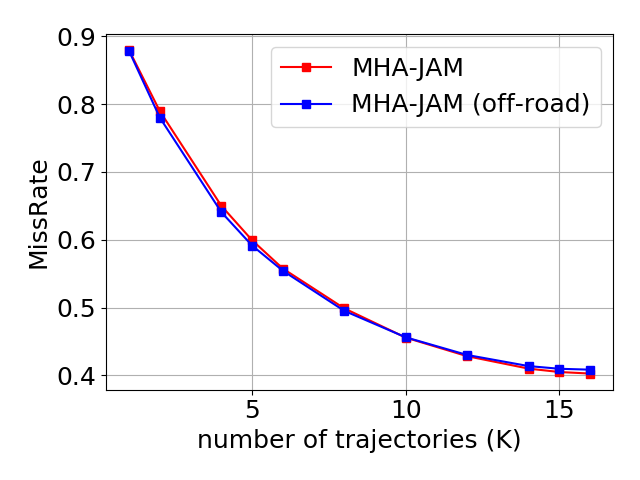}
\end{subfigure}
\begin{subfigure}[t]{0.3\textwidth}
\renewcommand\thesubfigure{i}
\includegraphics[width=0.9\linewidth]{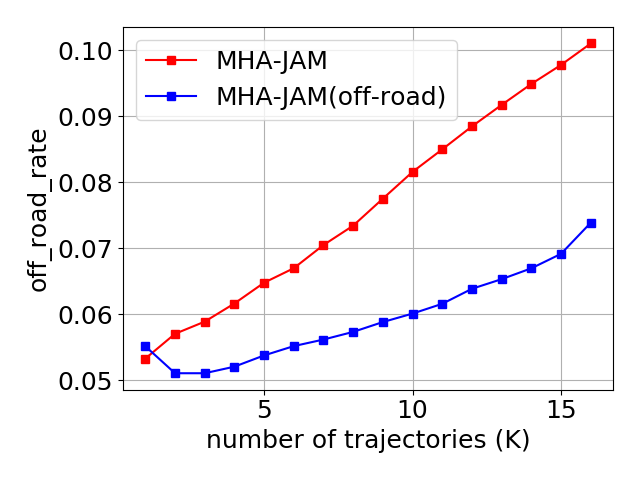}
\end{subfigure}
\caption{Effect of off-road loss}
\label{fig:loss0}
\end{subfigure}
\caption{\textbf{Ablation experiments:} We evaluate through ablation experiments, the importance of input cues (top), the effectiveness of a joint agent map representation for generating keys and values for attention heads (middle), the effectiveness of attention heads specialized for particular modes of the multimodal predictive distribution (middle), and finally the effectiveness of the auxiliary off-road loss (bottom). For each experiment we plot the metrics MinADE$_k$ (left), MissRate$_{k,2}$ (middle) and off-road rate (right) for the $k$ likeliest trajectories output by the models.}
\label{fig:Ablation}
\end{figure*}
To get a deeper insight on the relative contributions of various cues and modules affecting the overall performance, we perform the following ablation experiments. 

\vspace{0.05in}
\noindent \textbf{Importance of input cues:} Our model relies on two main inputs, past motion of surrounding agents and scene context captured with maps. To investigate the importance of each input, we compare our model MHA-JAM to two models: (1) MHA with purely agent inputs (MHA-A), and (2) MHA with purely map inputs (MHA-M). When considering only the surrounding agents without any information about the scene structure (MHA-A), the model shows poor results according to the three metrics MinADE$_k$, missRate$_{k,2}$, and off-road rate (cf. Figure~\ref{fig:Cues}). This highlights the importance of the map information to make more accurate and scene compliant predictions. Moreover, since MHA-JAM has the best performance, we infer that considering the surrounding agents also helps our model make a better prediction.


\vspace{0.05in}
\noindent\textbf{Advantage of using multiple attention heads:}
We train our model with different numbers of attention heads (L) and we compare the MinADE$_L$, MinFDE$_L$ and MissRate$_{L,2}$. Table~\ref{tab:diff_L} shows that all the metrics decrease when we increase the number of the attention heads. This proves the usefulness of using different attention heads to generate multimodal predictions.

\begin{table}[]
\centering
\caption{MinADE and MinFDE with different numbers of attention heads (L)}
\begin{tabular}{@{}lcccccc@{}}
\toprule
L        & 1    & 4    & 8    & 12   & 16   & 20   \\ \midrule
MinADE$_L$   & 3.48 & 1.72 & 1.26 & 1.13 & 1.02 & 1.00 \\
MinFDE$_L$   & 8.01 & 3.54 & 2.29 & 1.91 & 1.64 & 1.60 \\
MissRate$_{L,2}$ & 0.91 & 0.76 & 0.59 & 0.50 & 0.40 & 0.40 \\ \bottomrule
\end{tabular}
\label{tab:diff_L}
\end{table}

\begin{figure}[t]
\centering
\includegraphics[width=\columnwidth]{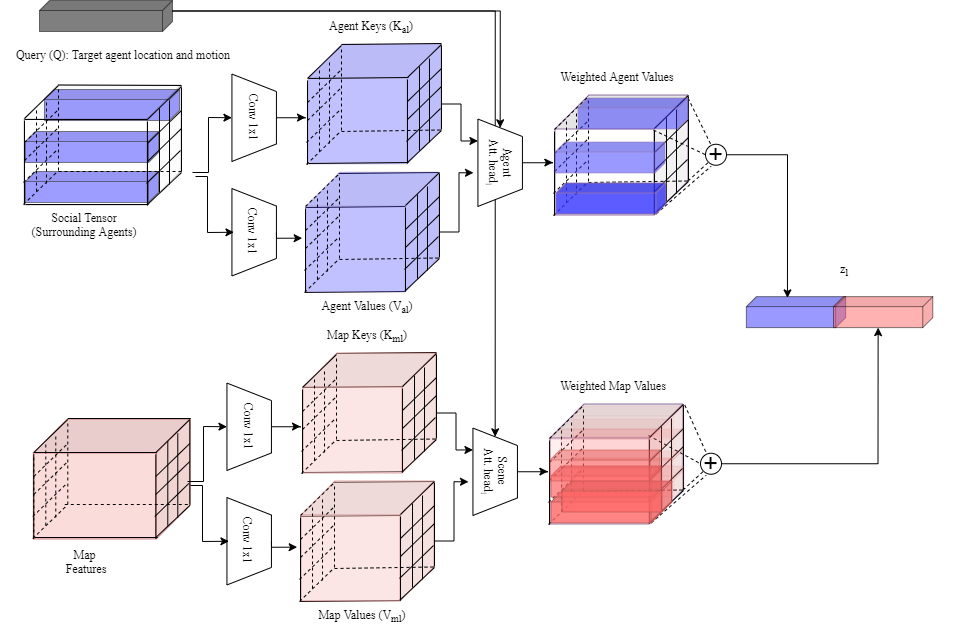}
\caption{\textbf{MHA with separate agent-map representation:} We compare our model to a baseline where attention weights are separately generated for the map and agent features}
\label{fig:SAM}
\end{figure}

\vspace{0.05in}
\noindent\textbf{Effectiveness of joint context representation:}
To prove the effectiveness of using joint context representation, we compare our model to MHA-SAM (MHA with Separate-Agent-Map representation). MHA-SAM is composed of two separate MHA blocks (cf. Figure~\ref{fig:SAM}): MHA on surrounding agents (similar to MHA-A) and on map (similar to MHA-M). We concatenate their outputs to feed them to the decoders. The main difference between MHA-SAM and MHA-JAM is that MHA-JAM generates keys and values in MHA using a joint representation of the map and agents while MHA-SAM computes keys and values of the map and agents features separately. Figure~\ref{fig:SAM_JAM} shows that  MHA-JAM performs better compared to MHA-SAM especially according to the off-road rate metric. This proves the benefit of applying attention on a joint spatio-temporal context representation composed of map and surrounding agents motion, over using separate attention blocks to model vehicle-map and vehicle-agents interaction independently.\\
\noindent\textbf{Effectiveness of a specialized attention heads:} We also compare our method to MHA-JAM (with Joint-Attention-Heads JAH). While MHA-JAM uses each attention head $head_l$ to generate a possible trajectory, MHA-JAM with joint attention heads uses a fully connected layer to combine the outputs of all attention heads $head_l, ~ l=1\dots L$. It generates each possible trajectory using a learnt combination of all the attention heads.

Comparing MHA-JAM and MHA-JAM (JAH) reveals that conditioning each possible trajectory on a context generated by one attention head performs better than generating each trajectory based on a combination of all attention heads.

\vspace{0.05in}
\noindent\textbf{Role of the Off-road loss:} Figure~\ref{fig:loss0} compares MHA-JAM trained with and without off-road loss (cf. Section~\ref{lossfn}). We notice that the off-road loss helps generating trajectories more compliant to the scene by reducing the off-road rate while maintaining good prediction precision.
\subsection{Qualitative Results}

\begin{figure*}[h]
\centering
\begin{subfigure}{\textwidth}
\centering
\begin{subfigure}[t]{0.25\textwidth}
\centering
\renewcommand\thesubfigure{i}
\includegraphics[height=4.2cm]{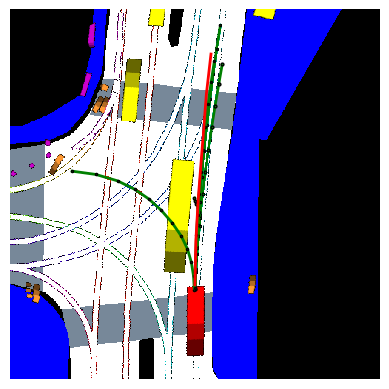}
\caption{Predicted trajectories}
\label{fig:Attn3_exp1_map}
\end{subfigure}
\hspace{0.01\textwidth}
\begin{subfigure}[t]{0.7\textwidth}
\centering
\addtocounter{subfigure}{-1}
\renewcommand\thesubfigure{ii}
\includegraphics[height=4.2cm]{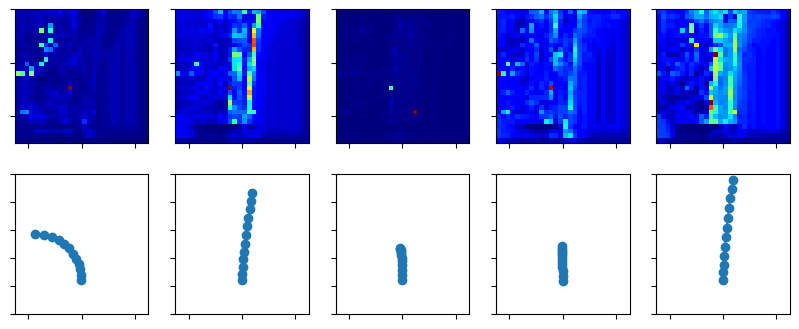}
\caption{5 most probable trajectories and their corresponding attention maps}
\label{fig:Attn3_exp1_head}
\end{subfigure}
    \addtocounter{subfigure}{-1}
   \caption{Example 1}
\label{fig:Attn3_exp1}
\end{subfigure}

\vspace{0.3cm}

\begin{subfigure}{\textwidth}
\centering
\begin{subfigure}[t]{0.25\textwidth}
\centering
\renewcommand\thesubfigure{i}
\includegraphics[height=4.2cm]{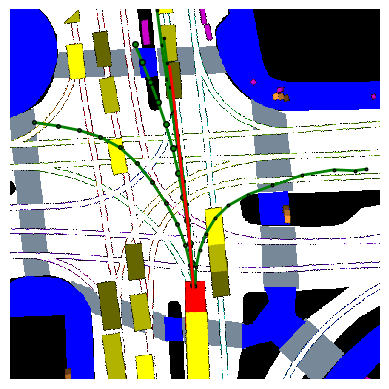}
\caption{Predicted trajectories}
\label{fig:Attn3_exp2_map}
\end{subfigure}
\hspace{0.01\textwidth}
\begin{subfigure}[t]{0.7\textwidth}
\centering
\addtocounter{subfigure}{-1}
\renewcommand\thesubfigure{ii}
\includegraphics[height=4.2cm]{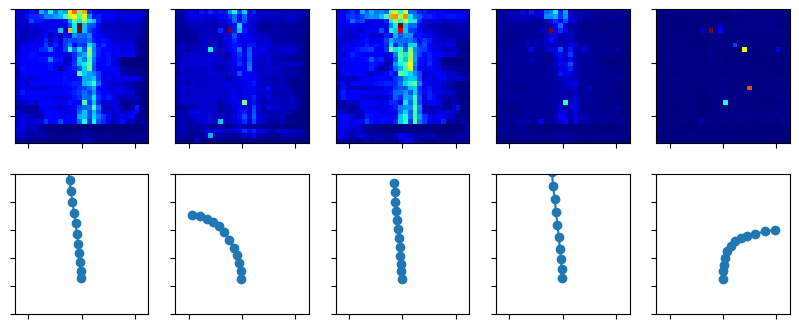}

\caption{5 most probable trajectories and their corresponding attention maps}
\label{fig:Attn3_exp2_head}
\end{subfigure}
\addtocounter{subfigure}{-1}

   \caption{Example 2}
\label{fig:Attn3_exp2}
\end{subfigure}

\caption{Examples of produced attention maps and trajectories with MHA-JAM (off-road) model }
\label{fig:Attn3}
\end{figure*}

\begin{figure*}[h]
\centering
\begin{subfigure}[t]{0.2\textwidth}
\includegraphics[width=0.85\textwidth]{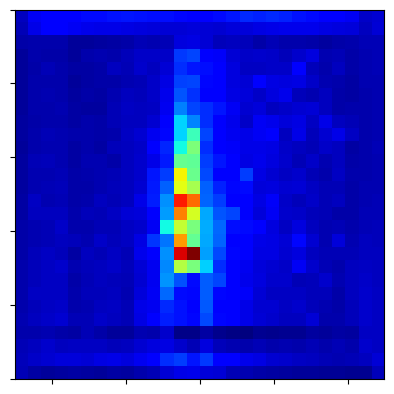}
\caption{Mean attention maps going straight (low speed)}
\label{fig:Attn_map_right}
\end{subfigure}
\hfill 
\begin{subfigure}[t]{0.2\textwidth}
\includegraphics[width=0.85\textwidth]{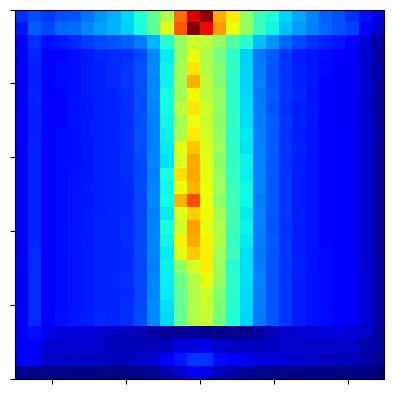}
\caption{Mean attention maps going straight (high speed)}
\label{fig:Attn_map_left}
\end{subfigure}
\hfill 
\begin{subfigure}[t]{0.2\textwidth}
\includegraphics[width=0.85\textwidth]{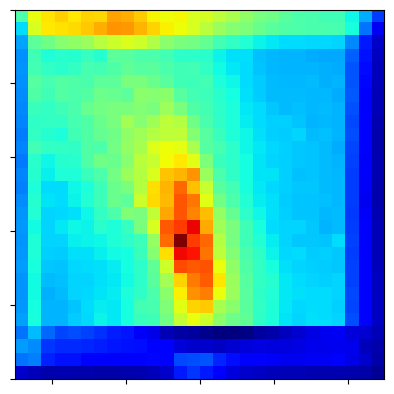}
\caption{Mean attention maps going left}
\label{fig:Attn_map_high_straight}
\end{subfigure}
\hfill 
\begin{subfigure}[t]{0.2\textwidth}
\includegraphics[width=0.85\textwidth]{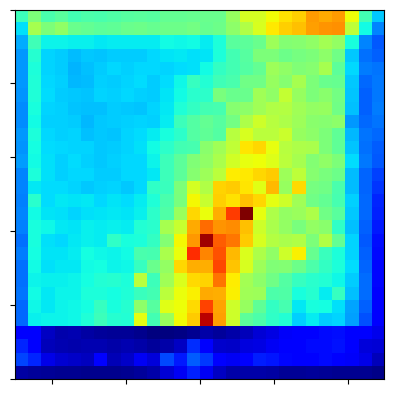}
\caption{Mean attention maps going right}
\label{fig:Attn_map_low_straight}
\end{subfigure}
   \caption{Visualisation of average attention maps over different generated maneuvers.}
\label{fig:Attn_map}
\end{figure*}

Figure~\ref{fig:Attn3} presents two examples of vehicle trajectory prediction, their corresponding 5 most probable generated trajectories and their associated attention maps. We notice that our proposed model MHA-JAM (off-road) successfully predicts diverse possible maneuvers; straight and left for the first  Example~\ref{fig:Attn3_exp1} and straight, left and right for the second Example~\ref{fig:Attn3_exp2}. In addition, it produces different attention maps which implies that it learnt to create specific context features for each predicted trajectories. For instance, the attention maps of the going straight trajectories, assign high weights to the drivable area in the straight direction and to the leading vehicles (the dark red cells). Moreover, They show focus on relatively close features when performed with low speed and further ones with high speed (cf. Example~\ref{fig:Attn3_exp1}).
For the left and right turns, in both examples, the corresponding attention maps seem to assign high weights to surrounding agents that could interact with the target vehicle while performing those maneuvers. For instance, in the left turn (cf. Example~\ref{fig:Attn3_exp2}), the attention map assigns high weights to vehicles in the opposite lane turning right. For the left turn of the first example and for the right turn of the second example, the attention maps assign high weights to pedestrians standing on both sides of the crosswalks. However, for the right turn, the model fails to take into account the traffic direction. 

Figure~\ref{fig:Attn_map} shows the average attention maps, for 4 generated possible maneuvers (going straight with low and high speed, left and right), over all samples in the test set. We note that each attention map assigns high weights, on average, to the leading vehicles, to surrounding agents and to the map cells in the direction of the performed maneuvers. This consolidates the previous observations in Figure~\ref{fig:Attn3}. We conclude that our model generates attention maps that focus on specific surrounding agents and scene features depending on the future possible trajectory.

\section{Concluding Remarks}\label{sec:conclusion}
This work tackled the task of vehicle trajectory prediction in an urban environment while considering interactions between the target vehicle, its surrounding agents and scene. To this end, we deployed a MHA-based method on a joint agents and map based global context representation. The model enabled each attention head to explicitly extract specific agents and scene features that help infer the driver's diverse possible behaviors. Furthermore, the visualisation of the attention maps reveals the importance of joint agents and map features and the interactions occurring during the execution of each possible maneuver. Experiments showed that our proposed approaches outperform the existing methods according to most of the metrics considered, especially the off-road metric. This highlights that the predicted trajectories comply with the scene structure. 


{\small
\bibliographystyle{ieee}
\bibliography{egbib}
}

\end{document}